\documentclass{article}
\usepackage{spconf,amssymb,amsmath,graphicx,epsfig,color,cite}
\usepackage[english,capitalize]{cleveref}
\usepackage{threeparttable}
\usepackage{comment}

\title{SIMULTANEOUS SPEECH RECOGNITION AND SPEAKER DIARIZATION FOR MONAURAL DIALOGUE RECORDINGS WITH TARGET-SPEAKER ACOUSTIC MODELS} 
\name{Naoyuki Kanda$^1$, Shota Horiguchi$^1$, Yusuke Fujita$^1$, Yawen Xue$^1$, Kenji Nagamatsu$^1$, Shinji Watanabe$^2$}
\address{$^1$Hitachi, Ltd., Japan\\ $^2$Johns Hopkins University, USA}
\newcommand{\argmax}{\mathop{\rm arg~max}\limits}

\begin{document}
\ninept
\maketitle
\begin{abstract}
This paper investigates the use of target-speaker automatic speech recognition (TS-ASR) for simultaneous speech recognition and speaker diarization of single-channel dialogue recordings. TS-ASR is a technique to automatically extract and recognize only the speech of a target speaker given a short sample utterance of that speaker. One obvious drawback of TS-ASR is that it cannot be used when the speakers in the recordings are unknown because it requires a sample of the target speakers in advance of decoding. To remove this limitation, we propose an iterative method, in which (i) the estimation of speaker embeddings and (ii) TS-ASR based on the estimated speaker embeddings are alternately executed. We evaluated the proposed method by using very challenging dialogue recordings in which the speaker overlap ratio was over 20\%. We confirmed that the proposed method significantly reduced both the word error rate (WER) and diarization error rate (DER). Our proposed method combined with i-vector speaker embeddings ultimately achieved a WER that differed by only 2.1 \% from that of TS-ASR given oracle speaker embeddings. Furthermore, our method can solve speaker diarization simultaneously as a by-product and achieved better DER than that of the conventional clustering-based speaker diarization method based on i-vector.
\end{abstract}
\begin{keywords}
multi-talker speech recognition, speaker diarization, deep learning
\end{keywords}
\section{Introduction}
\label{sec:intro}

Our main goal is to develop a monaural conversation transcription system 
that can not only perform automatic speech recognition (ASR) of multiple talkers 
but also determine who spoke the utterance when, known as speaker diarization \cite{tranter2006overview,fiscus2007rich}. 
For both ASR and  speaker diarization, the main difficulty comes from speaker overlaps. 
For example, a speaker-overlap ratio of about 15\% was reported in real meeting recordings \cite{yoshioka2018recognizing}. 
For such overlapped speech, neither conventional ASR nor  speaker diarization provides a result with sufficient accuracy. 
It is known that mixing two speech significantly degrades ASR accuracy \cite{yu2017recognizing,delcroix2018single,kanda2019two}.
In addition,
no speaker overlaps are assumed with most conventional  speaker diarization techniques, 
such as clustering of speech partitions (e.g. \cite{tranter2006overview,meignier2010lium,sell2014speaker,senoussaoui2013study,wang2018speaker}), 
which works only if there are no speaker overlaps.
Due to these difficulties, 
it is still very challenging to perform ASR and speaker diarization for monaural recordings of conversation.

One solution to the speaker-overlap problem is 
applying a speech-separation method such as deep clustering \cite{hershey2016deep} 
or deep attractor network \cite{chen2017deep}. However, a major drawback of such a method 
is that the training criteria for speech separation do not necessarily maximize 
the accuracy of the final target tasks. 
For example, if the goal is ASR, it will be better to use training criteria that directly maximize ASR accuracy.

In one line of research using ASR-based training criteria,
multi-speaker ASR based on permutation invariant training (PIT) has 
been proposed \cite{yu2017recognizing,chen2018progressive,settle2018end,seki2018purely,chang2019end}. 
With PIT, the label-permutation problem is solved by considering all possible permutations 
when calculating the loss function \cite{yu2017permutation}. 
PIT was first proposed for speech separation \cite{yu2017permutation} and 
soon extended to ASR loss with promising results \cite{yu2017recognizing,chen2018progressive,settle2018end,seki2018purely,chang2019end}. 
However, a PIT-ASR model produces transcriptions {\it for each utterance} of speakers in an unordered manner, 
and it is no longer straightforward to solve speaker permutations {\it across utterances}. To make things worse, a PIT model trained with ASR-based loss normally does not produce separated speech waveforms, which makes speaker tracing more difficult. 

In another line of research, target-speaker (TS) ASR, which automatically extracts and transcribes 
only the target speaker's utterances given a short sample of that speaker's speech, 
has been proposed \cite{zmolikova2017speaker,delcroix2018single}. 
{\v{Z}}mol{\'\i}kov{\'a} et al. proposed a target-speaker neural beamformer 
that extracts a target speaker's utterances given a short sample 
of that speaker's speech \cite{zmolikova2017speaker}. 
This model was recently extended to handle ASR-based loss to maximize ASR accuracy 
with promising results \cite{delcroix2018single}. 
TS-ASR can naturally solve the speaker-permutation problem {\it across utterances}. 
Importantly, if we can execute TS-ASR for each speaker correctly,  speaker diarization is solved at the same time 
just by extracting the start and end time information of the TS-ASR result. 
However, one obvious drawback of TS-ASR is that it cannot be applied 
when the speakers in the recordings are unknown because it requires a sample of the target speakers in advance of decoding.

Based on this background, we propose a speech recognition and  speaker diarization method 
that is based on TS-ASR but can be applied without knowing the speaker information in advance. 
To remove the limitation of TS-ASR, 
we propose an iterative method, in which
(i) the estimation of target-speaker embeddings and (ii) TS-ASR based on the estimated embeddings 
are alternately executed. 
As an initial trial, we evaluated the proposed method 
by using real dialogue recordings in the Corpus of Spontaneous Japanese (CSJ). 
Although it contains the speech of only two speakers, 
the speaker-overlap ratio of the dialogue speech is very high; 20.1\% . 
Thus, this is very challenging even for state-of-the-art ASR and  speaker diarization.
We show that the proposed method effectively reduced both word error rate (WER) and diarizaton error rate (DER).

\begin{figure*}[t]
	\centering
	\includegraphics[width=165mm]{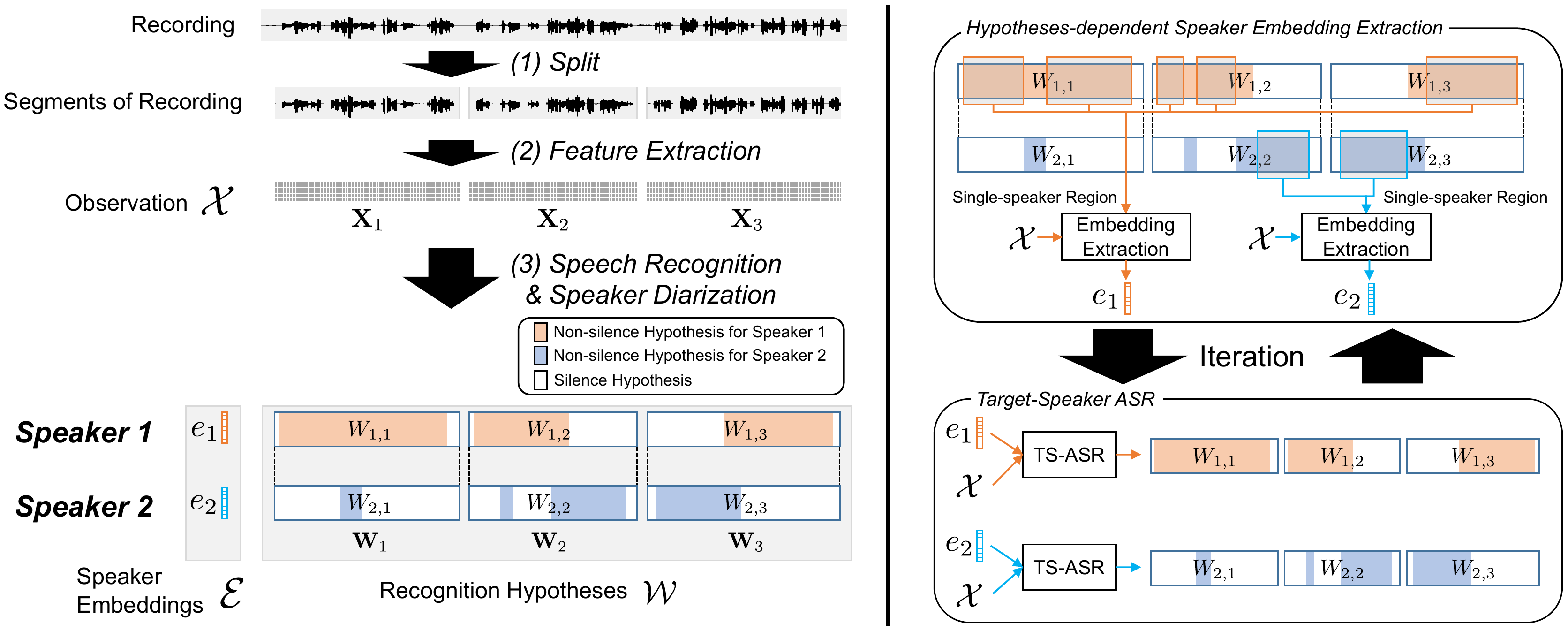}
	\vspace{-4mm}
	\caption{{\it (left) Overview of simultaneous speech recognition and speaker diarization, (right) proposed iterative maximization method.}}
	\label{fig:overview}
	\vspace{-3mm}
\end{figure*}

\section{Simultaneous ASR and  Speaker Diarization}
\label{sec:proposed}

In this section, we first explain the problem we targeted 
then the proposed method with reference to Figure \ref{fig:overview}.

\subsection{Problem statement}

The overview of the problem 
is shown in Figure \ref{fig:overview} (left).
We assume a sequence of observations $\mathcal{X}=\{{\bf X}_1,...,{\bf X}_U\}$, 
where $U$ is the number of observations, and ${\bf X}_u$ is the $u$-th observation consisting 
of a sequence of acoustic features. 
Such a sequence is naturally generated when we separate a long recording into small segments 
based on 
voice activity detection
which is a basic preprocess for ASR so as not to generate 
overly large lattices. 
We also assume a tuple of word hypotheses ${\bf W}_u=(W_{1,u},...,W_{J,u})$ for an observation ${\bf X}_u$ 
where $J$ is the number of speakers, and $W_{j,u}$ represents the speech-recognition hypothesis 
of the $j$-th speaker given observation ${\bf X}_u$. 
We assume $W_{j,u}$ contains not only word sequences but also 
their corresponding
frame-level time alignments of phonemes and silences. 
Finally, we assume a tuple of speaker embeddings $\mathcal{E}=(e_1, ..., e_J)$, 
where $e_j\in \mathbb{R}^d$ represents the $d$-dim speaker embedding of the $j$-th speaker.

Then, our objective is to find 
the best possible $\mathcal{W}=\{{\bf W}_1,...,{\bf W}_U\}$ given a sequence of observations $\mathcal{X}$ as follows.
\begin{align}
\tilde{\mathcal{W}} &= \argmax_{\mathcal{W}} P(\mathcal{W}|\mathcal{X}) \label{eq:first} \\
 &= \argmax_{\mathcal{W}} \{\sum_{\mathcal{E}} P(\mathcal{W},\mathcal{E}|\mathcal{X})\} \label{eq:second}\\
 &\simeq \argmax_{\mathcal{W}} \{\max_{\mathcal{E}} P(\mathcal{W},\mathcal{E}|\mathcal{X})\} \label{eq:third}
\end{align}
Here, the starting point  is the conventional maximum a posteriori-based decoding 
given $\mathcal{X}$ but 
for multiple speakers.
We then introduce the speaker embeddings $\mathcal{E}$ 
as a hidden variable (Eq. \ref{eq:second}).
Finally, we approximate the summation by using a max operation (Eq. \ref{eq:third}).

Our motivation to introduce 
$\mathcal{E}$, which is constant across all observation indices $u$, is to 
explicitly enforce the order of speakers in $\mathcal{W}$ to be constant over indices $u$.
It should be emphasized that if we can solve the problem,  speaker diarization is solved at the same time 
just by extracting the start and end time information of each hypothesis in $\mathcal{W}$. 
Also note that there are $J!$ possible solutions 
by swapping the order of speakers in $\mathcal{E}$, and it is sufficient to find just one such solution.

\subsection{Iterative maximization}
\label{sec:iterative}

It is not easy to directly solve $P(\mathcal{W},\mathcal{E}|\mathcal{X})$, 
so we propose to alternately maximize $\mathcal{W}$ and $\mathcal{E}$.
Namely, we first fix $\underline{\mathcal{W}}$ and find $\mathcal{E}$
that maximizes 
$P(\underline{\mathcal{W}},\mathcal{E}|\mathcal{X})$.
We then fix $\underline{\mathcal{E}}$ and find  $\mathcal{W}$
that maximizes 
$P(\mathcal{W},\underline{\mathcal{E}}|\mathcal{X})$.
By iterating this procedure,
$P(\mathcal{W},\mathcal{E}|\mathcal{X})$ can be increased monotonically.
Note that 
it can be said by a simple application of the chain rule that
finding $\mathcal{E}$
that maximizes $P(\underline{\mathcal{W}},\mathcal{E}|\mathcal{X})$ with a fixed
$\underline{\mathcal{W}}$  is equivalent to 
finding $\mathcal{E}$
that maximizes $P(\mathcal{E}|\underline{\mathcal{W}},\mathcal{X})$.
The same thing can be said 
for the estimation of 
$\mathcal{W}$
with a fixed $\underline{\mathcal{E}}$. 

For the $(i)$-th iteration of the maximization ($i\in\mathbb{Z}^{\geq 0}$), 
we first find the most plausible estimation of $\mathcal{E}$ 
given the $(i-1)$-th speech-recognition hypothesis $\tilde{\mathcal{W}}^{(i-1)}$ as follows.
\begin{align}
\tilde{\mathcal{E}}^{(i)}&=\begin{cases}
\argmax_{\mathcal{E}} P(\mathcal{E}|\tilde{\mathcal{W}}^{(i-1)},\mathcal{X}) & (i \geq 1) \label{eq:emb1}\\
\argmax_{\mathcal{E}} P(\mathcal{E}|\mathcal{X}) & (i=0)
\end{cases}
\end{align}
Here, the estimation of $\tilde{\mathcal{E}}^{(i)}$ is dependent on $\tilde{\mathcal{W}}^{(i-1)}$ 
for $i \geq 1$.
Assume that the overlapped speech corresponds to a ``third person'' who is different from any person in the recording, Eq. \ref{eq:emb1} can be achieved by estimating
the speaker embeddings only from non-overlapped regions (upper part of Figure \ref{fig:overview} (right)).
In this study, 
we used i-vector \cite{dehak2011front} as the representation of speaker embeddings, 
and estimated i-vector based only
on the non-overlapped region given $\tilde{\mathcal{W}}^{(i-1)}$ for each speaker\footnote{The idea to extract speaker embeddings from non-overlapped regions has been proposed (e.g. \cite{kanda2018hitachi,manohar2019acoustic})}. 
Note that, since we do not have an estimation of $\mathcal{W}$ for the first iteration, $\tilde{\mathcal{E}}^{(0)}$ is 
initialized only by $\mathcal{X}$. 
In this study, we 
estimated the i-vector for each speaker given
 the speech region that was estimated by the clustering-based speaker diarization method.
More precicely,
we estimated the i-vector for each ${\bf X}_u$ then applied $J$-cluster K-means clustering. 
The center of each cluster\footnote{Using cluster centers does not strictly follow Eq. \ref{eq:emb1}, but we 
used them for the procedural simplicity.} was used for the initial 
speaker embeddings $\tilde{\mathcal{E}}^{(0)}$.

We then update 
$\mathcal{W}$ given speaker embeddings $\tilde{\mathcal{E}}^{(i)}$.
\begin{align}
\tilde{\mathcal{W}}^{(i)}&=\argmax_{\mathcal{W}} P(\mathcal{W}|\tilde{\mathcal{E}}^{(i)},\mathcal{X}) \label{eq:w-update1}\\
 &\simeq \argmax_{{\bf W}_1,...,{\bf W}_U} \prod_u P({\bf W}_u|\tilde{\mathcal{E}}^{(i)},{\bf X}_u) \label{eq:w-update2} \\
 &\simeq \argmax_{{\bf W}_1,...,{\bf W}_U} \prod_u \prod_j P(W_{j,u}|\tilde{e}_j^{(i)},{\bf X}_u) \label{eq:w-update}
\end{align}
Here, we estimate the most plausible hypotheses $\mathcal{W}$ given estimated embeddings $\tilde{\mathcal{E}}^{(i)}$ 
and observation $\mathcal{X}$ (Eq. \ref{eq:w-update1}).
We then assume the conditional independence  of ${\bf W}_u$ given ${\bf X}_u$ for each segment $u$ (Eq. \ref{eq:w-update2}).
Finally, we further assume  the conditional independence of $W_{j,u}$ given $\tilde{e}_j^{(i)}$ for each speaker $j$ (Eq. \ref{eq:w-update}).
The final equation can be solved by applying TS-ASR for each segment $u$ for each speaker $j$ (lower part of Figure \ref{fig:overview} (right)).
We will review the detail of TS-ASR in the next section.

\section{TS-ASR: Review}
\label{sec:ts-asr}

\subsection{Overview of TS-ASR}

TS-ASR is a technique to extract and recognize only the speech of a target speaker given 
a short sample utterance of that speaker \cite{zmolikova2017speaker,vzmolikova2017learning,delcroix2018single}. 
Originally, the sample utterance was fed into a special neural network that 
outputs an averaged embedding to control the weighting 
of speaker-dependent blocks of the acoustic model (AM). 
However, to make the problem simpler, 
we assume that a $d$-dimensional speaker embedding $e_{\rm tgt}\in \mathbb{R}^d$ is 
extracted from the sample utterance.
In this context, TS-ASR can be expressed 
as the problem to find the best hypothesis $W_{\rm tgt}$ given 
observation ${\bf X}$ and speaker embedding $e_{\rm tgt}$ as follows.
\begin{align}
\tilde{W}_{\rm tgt}=\argmax_{W_{\rm tgt}} P(W_{\rm tgt}|e_{\rm tgt},{\bf X}) \label{eq:ts-asr}
\end{align}
If we have a well-trained TS-ASR, Eq. \ref{eq:w-update} can be solved by simply
applying the TS-ASR for each segment $u$ for each speaker $j$.

\subsection{TS-AM with auxiliary output network}
\subsubsection{Overview}

\begin{figure}[t]
	\centering
	\includegraphics[width=80mm]{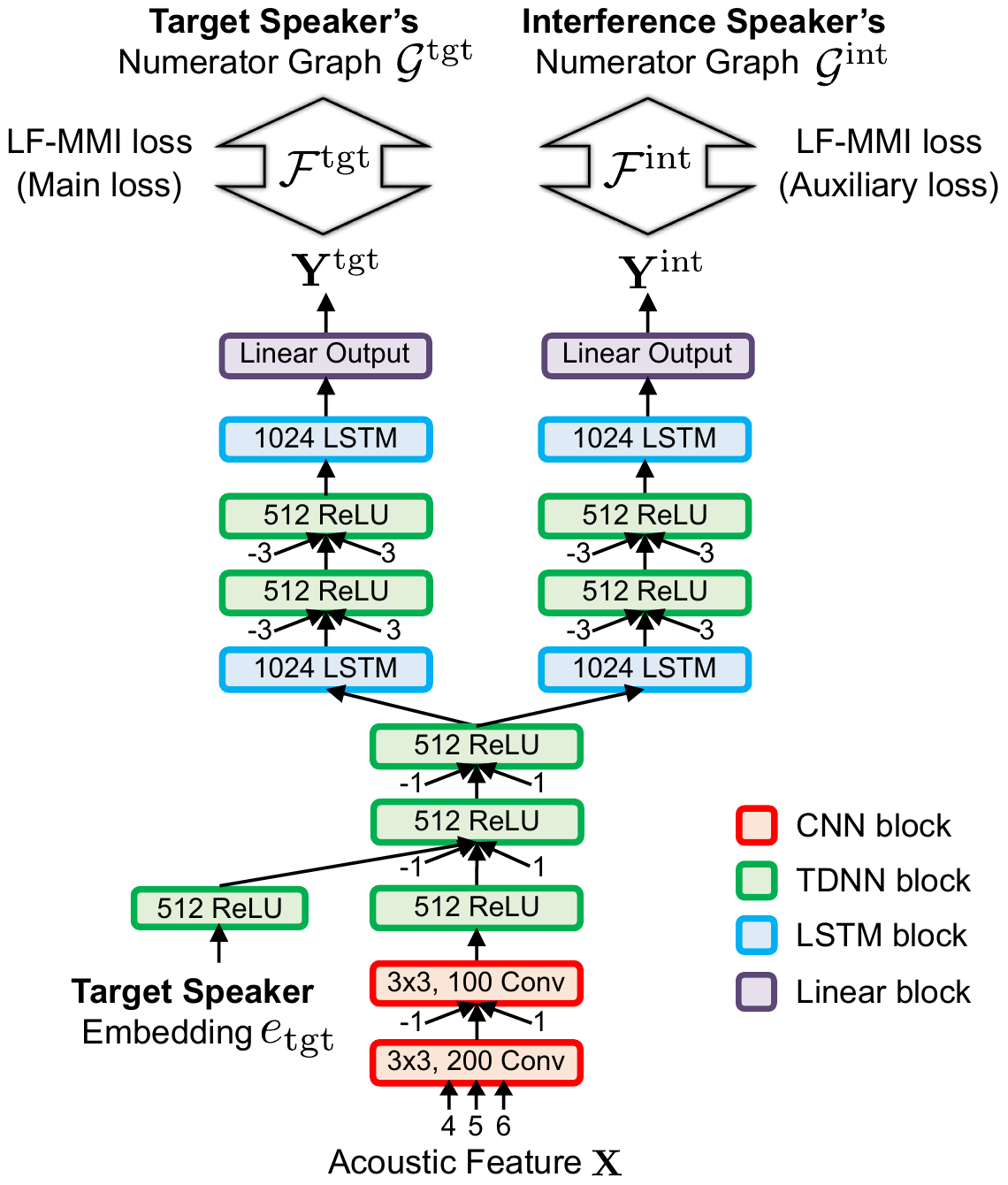}
	\vspace{-3mm}
	\caption{\textit{Overview of target-speaker AM architecture with auxiliary interference speaker loss \cite{kanda2019two}. A number with an arrow indicates a time splicing index, which forms the basis of a time-delay neural network (TDNN) \cite{peddinti2015time}. The input features were advanced by five frames, which has the same effect as reference label delay.}}
	\label{fig:ts-am}
	\vspace{-2mm}
\end{figure}

Although any speech recognition architecture can be used for TS-ASR, 
we adopted a variant of the TS-AM that was recently proposed 
and has promising accuracy \cite{kanda2019two}. Figure \ref{fig:ts-am} describes the TS-AM 
that we applied for this study. 
This model has two input branches. 
One branch accepts acoustic features ${\bf X}$ as a normal AM 
while the other branch accepts an embedding $e_{\rm tgt}$ that 
represents the characteristics of the target speaker. 
In this study, we used a log Mel-filterbank (FBANK) and i-vector \cite{dehak2011front,saon2013speaker} 
for the acoustic features and target-speaker embedding, respectively. 

A unique component of the model is in its output branch. 
The model has multiple output branches that produce outputs ${\bf Y}^{\rm tgt}$ and ${\bf Y}^{\rm int}$ 
for the loss functions for the target and interference speakers, respectively. 
The loss for the target speaker is defined to 
maximize the target-speaker ASR accuracy, 
while the loss for interference speakers is defined to
 maximize the interference-speaker ASR accuracy. 
We used lattice-free maximum mutual information (LF-MMI) \cite{povey2016purely} for both criteria. 

The original motivation of the output branch for interference speakers was 
the improvement of TS-ASR by achieving
a better representation for speaker separation in the shared layers.
However, it was also shown that 
the output branch for interference speakers can be used 
for the secondary ASR for interference speakers given 
the embedding of the target speaker \cite{kanda2019two}.
In this paper, we found out that 
the latter property worked very well for the ASR for dialogue recordings,
which will be explained in the evaluation section.

The network is trained with a mixture of multi-speaker speech 
given their transcriptions. 
We assume that, for each training sample, 
(a) transcriptions of at least two speakers are given, 
(b) the transcription for the target speaker is marked 
so that we can identify the target speaker's transcription, 
and (c) a sample for the target speaker can be used to extract speaker embeddings. 
These assumptions can be easily satisfied by artificially generating training data by mixing the speech of multiple speakers.

\subsubsection{Loss function}
The main loss function for the target speaker is defined as 
\begin{align}
\mathcal{F}^{\rm tgt}&=\sum_{u} {\mathrm{LFMMI}}({\bf Y}_{u}^{\rm tgt};\mathcal{G}^{\rm tgt}_{u}),  \\
=&\sum_{u} \sum_{{\bf S}} -P({\bf S}|{\bf Y}_{u}^{\rm tgt},\mathcal{G}^{\rm tgt}_{u}) \log P({\bf S}|{\bf Y}_{u}^{\rm tgt},\mathcal{G}^{D}), \label{eq:mmi}
\end{align}
where $u$ corresponds to the index of training samples in this case. 
The term $\mathcal{G}^{\rm tgt}_u$ indicates a numerator (or reference) graph 
that represents a set of possible correct state sequences for the utterance of the target speaker 
of the $u$-th training sample, ${\bf S}$ denotes a hypothesis state sequence 
for the $u$-th training sample, and $\mathcal{G}^{D}$ denotes a denominator graph, 
which represents a possible hypothesis space and normally consists of a 4-gram phone language model 
in LF-MMI training \cite{povey2016purely}.

The auxiliary interference speaker loss is then defined to maximize the interference-speaker ASR accuracy, 
which we expect to enhance the speaker separation ability of the neural network. This loss is defined as
\begin{align}
\mathcal{F}^{\rm int}=\sum_{u} {\mathrm{LFMMI}}({\bf Y}_{u}^{\rm int};\mathcal{G}^{\rm int}_{u}), 
\end{align}
where $\mathcal{G}^{\rm int}_u$ denotes a numerator (or reference) graph 
that represents a set of possible correct state sequences 
for the utterance of the interference speaker of the $u$-th training sample.

Finally, the loss function $\mathcal{F}^{\rm comb}$ for training is 
defined as the combination of the target and interference losses,
\begin{align}
\mathcal{F}^{\rm comb}=\mathcal{F}^{\rm tgt}+\alpha\cdot\mathcal{F}^{\rm int}, 
\end{align}
where $\alpha$ is the scaling factor for the auxiliary loss. 
In our evaluation, we set $\alpha=1.0$. Setting $\alpha=0.0$, however, corresponds to normal TS-ASR.

\section{Evaluation}
\label{sec:eval}
\subsection{Experimental settings}
\label{sec:settings}

\subsubsection{Main evaluation data: real dialogue recordings}
We conducted our experiments on the CSJ \cite{maekawa2003corpus}, 
which is one of the most widely used evaluation sets for Japanese speech recognition. 
The CSJ consists of more than 600 hrs of Japanese recordings.

While most of the content is lecture recordings by a single speaker, 
CSJ also contains 11.5 hrs of 
54 dialogue recordings\footnote{We excluded 4 out of 58 dialogue recordings that have speaker duplication with the official E1, E2, and E3 evaluation sets.} 
(average 12.8 min per recording) 
with two speakers, 
which were the main target of ASR and  speaker diarization in this study. 
During the dialogue recordings, two speakers sat in two adjacent sound proof chambers divided by a glass window.
They could talk with each other over voice connection through 
a headset for each speaker. 
Therefore, speech was recorded separately for each speaker, 
and we generated mixed monaural recordings by mixing the corresponding speeches of two speakers. 
When mixing two recordings, we did not apply any normalization of speech volume. 
Due to this recording procedure, we were able to use non-overlapped speech to evaluate the oracle WERs.

It should be noted that, although the dialogue consisted of only two speakers, 
the speaker overlap ratio of the recordings was very high due to many backchannels and natural turn-taking. 
Among all recordings, 16.7\% of the region was overlapped speech while 66.4\% was spoken by a single speaker. 
The remaining 16.9\% was silence. 
Therefore, 20.1\% (=16.7/(16.7+66.4)) of speech regions was speaker overlap. 
From the viewpoint of ASR, 33.5\% (= (16.7*2)/(16.7*2+66.4)) of the total duration
 to be recognized was overlapped. 
These values were even higher than those reported for meetings with more than 
two speakers \cite{ccetin2006analysis,yoshioka2018recognizing}. 
Therefore, these dialogue recordings are very challenging for both ASR and  speaker diarization. 
We observed significantly high WER and DER, which is discussed in the next section.

\subsubsection{Sub evaluation data: simulated 2-speaker mixture}
To evaluate TS-ASR, we also used 
the simulated 2-speaker-mixed data by mixing
the three official single-speaker evaluation sets of CSJ, i.e., E1, E2, and E3 \cite{kawahara2003benchmark}. 
Each set includes different groups of 10 lectures (5.6 hrs, 30 lectures in total). 
The E1 set consists of 10 lectures of 10 male speakers, 
and E2 and E3 each consists of 10 lectures of 5 female and 5 male speakers. 
We generate two-speaker mixed speech by adding randomly selected speech (= interference-speaker speech) 
to the original speech (= target-speaker speech) with 
the constraint that the target and interference speakers were different, 
and each interference speaker was selected only once from the dataset. 
When we mixed the two speeches, 
we configured them to have the same power level, 
and shorter speech was mixed with the longer speech from a random starting point 
selected to ensure the end point of the shorter one did not exceed that of the longer one.

\subsubsection{Training data and training settings}
The rest of the 571 hrs of 3,207 lecture recordings 
(excluding the same speaker's lectures in the evaluation sets) were used for AM and language model (LM) training. 
We generated two-speaker mixed speech for training data in accordance with the following protocol.
\begin{enumerate}
\setlength{\parskip}{0cm} 
\setlength{\itemsep}{0cm} 
\item Prepare a list of speech samples (= main list).
\item Shuffle the main list to create a second list under the constraint that the same speaker does not appear in the same line in the main and second lists.
\item Mix the audio in the main and second lists one-by-one with a specific signal-to-interference ratio (SIR). For training data, we randomly sampled an SIR as follows.
\begin{itemize}
\item In 1/3 probability, sample the SIR from a uniform distribution between -10 and 10 dB. 
\item In 1/3 probability, sample the SIR from a uniform distribution between 10 and 60 dB. The transcription of the interference speaker was set to null.
\item In 1/3 probability, sample the SIR from a uniform distribution between -60 and -10 dB. The transcription of the target speaker was set to null.
\end{itemize}
\item The volume of each mixed speech was randomly changed to enhance robustness against volume difference.
\end{enumerate}
A speech for extracting a speaker embedding was also randomly selected for each speech mixture from the main list. Note that the random perturbation of volume was applied only for the training data, not for evaluation data.

We trained a TS-AM consisting of a convolutional neural network (CNN), 
time-delay NN (TDNN) \cite{waibel1989phoneme}, 
and long short-term memory (LSTM) \cite{hochreiter1997long}, as shown in \cref{fig:ts-am}. 
The input acoustic feature for the network was a 40-dimensional FBANK without normalization. 
A 100-dimensional i-vector was also extracted and used for the target-speaker embedding 
to indicate the target speaker. 
For extracting this i-vector, we randomly selected an utterance of the same speaker. 
We conducted 8 epochs of training on the basis of LF-MMI, 
where the initial learning rate was set to 0.001 and exponentially decayed to 0.0001 by the end of the training. 
We applied $l2$-regularization and CE-regularization \cite{povey2016purely} 
with scales of 0.00005 and 0.1, respectively. 
The leaky hidden Markov model coefficient was set to 0.1. 
A backstitch technique \cite{wang2017backstitch} with a backstitch scale of 1.0 and backstitch interval of 4 was also used. 

For comparison, we trained another TS-AM without the auxiliary loss.
We also trained a ``clean AM'' using clean, non-speaker-mixed speech. 
For this clean model, we used a model architecture without the auxiliary output branch, 
and an i-vector was extracted every 100 msec for online speaker/environment adaptation. 

In decoding, we used a 4-gram LM trained using the transcription of the training data. 
All our experiments were conducted on the basis of the Kaldi toolkit \cite{povey2011kaldi}.

\subsection{Preliminary experiment with simulated 2-speaker mixture}
\subsubsection{Evaluation of TS-ASR}

We first evaluated the TS-AM 
with two-speaker mixture of the E1, E2, and E3 evaluation sets. 
For each test utterance, a sample of the target speaker was randomly selected from the other utterances 
in the test set. 
We used the same random seed over all experiments, so that they could be conducted under the same conditions.

\begin{table}[t]
\caption{\label{tab:first} {\it WERs (\%) for two-speaker-mixed evaluation sets of CSJ.}}
\centering
{\footnotesize
\begin{tabular}{cc|ccc|c} \hline
Model              & Evaluation Data     & E1 & E2 & E3 & Avg. \\  \hline
Clean AM           & 1-spk.                & 8.94  & 7.31  & 7.44  &  7.90 \\
Clean AM           & 2-spk. mixed        & 87.60 & 85.44 & 91.05 &  88.03\\
TS-AM ($\alpha=0.0$) & 2-spk. mixed      & 26.01 & 18.16 & 18.16 & 20.78 \\
TS-AM ($\alpha=1.0$) & 2-spk. mixed      & {\bf 25.68} & {\bf 17.94} & {\bf 17.96} & {\bf 20.53} \\ \hline
\end{tabular}
}
\vspace{-3mm}
\end{table}

The results are listed in Table \ref{tab:first}. 
Although the clean AM produced a WER of 7.90\% for the original clean dataset, 
the WER severely degraded to 88.03\% by mixing two speakers. 
The TS-AM then significantly recovered the WER to 20.78\% ($\alpha=0.0$).
Although the improvement was not so significant 
compared with that reported in \cite{kanda2019two},
the auxiliary loss further improved the WER to 20.53\% ($\alpha=1.0$). 
Note that E1 contains only male speakers while E2 and E3 contain both female and male speakers. 
Because of this, 
E1 showed larger degradation of WER when 2 speakers were mixed.

\begin{table}[t]
\caption{\label{tab:second} {\it WERs (\%) for two-speaker-mixed evaluation sets of CSJ. Main output branch was used for target-speaker ASR and auxiliary output branch was used for interference-speaker ASR.}}
\centering
\begin{tabular}{c|cc} \hline
Test set               & Target spk. & Interference spk.  \\ \hline
E1 (10 male)       & 25.68 & 26.91 \\
E2 (5 female, 5 male)      & 17.94 & 18.46 \\
E3 (5 female, 5 male)      & 17.96 & 18.36 \\ \hline
Avg. & 20.53 & 21.24 \\ \hline
\end{tabular}
\vspace{-3mm}
\end{table}

\begin{table*}[t]
\caption{\label{tab:dialogue} {\it WERs (\%) for dialogue speech in CSJ}}
\centering
\begin{threeparttable}
\begin{tabular}{c|cccc|cc|c} \hline
\# &\multicolumn{2}{c}{Speaker Embeddings} & AM        & Evaluation  & \multicolumn{2}{c|}{Gender Pair} & Total \\
& Initialization   & Update   &                                     & Data      & Different & Same &  \\  \hline
1 &-                &  -       & Clean-AM                            & 1-spk.             & 18.49\tnote{$\dagger$} & 21.14\tnote{$\dagger$} & 19.93\tnote{$\dagger$} \\ \hline
2 &Oracle           &  -       & Clean-AM w/ ${\bf e}_1$ \& Clean-AM w/ ${\bf e}_2$ & 2-spk. mixed             & 94.46\tnote{$\dagger$}  &   94.01\tnote{$\dagger$}     &  94.22\tnote{$\dagger$}\\
3 &Oracle           &  -       & TS-AM (tgt) w/ ${\bf e}_1$ \& TS-AM (tgt) w/ ${\bf e}_2$ & 2-spk. mixed & 26.83\tnote{$\dagger$} & 47.33\tnote{$\dagger$} & 37.96\tnote{$\dagger$} \\
4 &Oracle           &  -       & TS-AM (tgt) w/ ${\bf e}_1$ \& TS-AM (int) w/ ${\bf e}_1$ & 2-spk. mixed & 25.99\tnote{$\dagger$} & 53.80\tnote{$\dagger$} & 41.09\tnote{$\dagger$} \\ \hline
5 &K-means          &  ${(i=0)}$       & TS-AM (tgt) w/ ${\bf e}_1$ \& TS-AM (tgt) w/ ${\bf e}_2$ & 2-spk. mixed & 40.99 & 64.97 & 54.01 \\
6 &K-means          &  ${(i=0)}$       & TS-AM (tgt) w/ ${\bf e}_1$ \& TS-AM (int) w/ ${\bf e}_1$ & 2-spk. mixed & 30.00 & 58.61 & 45.54 \\ 
7 &K-means          &  $i=1$  & TS-AM (tgt) w/ ${\bf e}_1$ \& TS-AM (int) w/ ${\bf e}_1$ & 2-spk. mixed & 26.45 & 53.93 & 41.37 \\
8 &K-means          &  $i=2$  & TS-AM (tgt) w/ ${\bf e}_1$ \& TS-AM (int) w/ ${\bf e}_1$ & 2-spk. mixed & 25.46 & 52.82 & 40.31 \\
9 &K-means          &  $i=3$  & TS-AM (tgt) w/ ${\bf e}_1$ \& TS-AM (int) w/ ${\bf e}_1$ & 2-spk. mixed & {\bf 25.20} & {\bf 52.50} & {\bf 40.03} \\ \hline
\end{tabular}
\begin{tablenotes}
	\item[$\dagger$] Result obtained with some oracle information such as non-overlapped evaluation data or oracle speaker embeddings
\end{tablenotes}
\end{threeparttable}
\vspace{-5mm}
\end{table*}

\begin{table}[t]
\caption{\label{tab:der} {\it DERs (\%) for dialogue speech}}
\centering
\begin{threeparttable}
\begin{tabular}{cccc} \hline
Method               & \multicolumn{2}{c}{Gender Pair} & Total \\ 
                       & Different & Same &  \\ \hline
i-vector with K-means                           & 25.94 & 37.32     & 32.37 \\ 
\# 6 of Table \ref{tab:dialogue}        & 15.99 & 37.00    & 27.87 \\
\# 9 of Table \ref{tab:dialogue}        & {\bf 10.76} & {\bf 35.30}    & {\bf 24.63} \\ \hline \hline
i-vector with AHC \cite{sell2018diarization}\tnote{$\ddagger$} & 14.34 & 38.48  &  27.99\\ 
x-vector with AHC \cite{sell2018diarization}\tnote{$\ddagger$}& 13.77 & 30.02  &  22.96\\ \hline
\end{tabular}
\begin{tablenotes}
	{\scriptsize \item[$\ddagger$] Trained using combination of Switchboard and NIST SRE datasets}
\end{tablenotes}
\end{threeparttable}
\vspace{-5mm}
\end{table}

\begin{table}[t]
\caption{\label{tab:der-detail} {\it Details of DER (\%) for \# 9 of Table \ref{tab:dialogue}}}
\centering
{\footnotesize
\begin{tabular}{c|ccc|c} \hline
                       & Miss & False Alarm & Confusion & DER \\ \hline
Different Gender Pair  & 9.6 & 0.8 & 0.4  & 10.76 \\
Same Gender Pair       & 22.5 & 2.2  & 10.6 & 35.30 \\ \hline
Total      & 16.9 & 1.6  & 6.2 & 24.63 \\ \hline
\end{tabular}
}
\vspace{-3mm}
\end{table}

\subsubsection{Interference-speaker ASR by auxiliary output branch}
Before moving to the evaluation of dialogue recordings, 
we also evaluated the use case of the auxiliary output branch for interference speakers 
to conduct interference-speaker ASR. 
In this experiment, we provided the target speaker's embeddings for the TS-AM
and evaluated the WERs of the ASR results 
using the auxiliary output branch. 
The results are shown in \cref{tab:second}. 
We confirmed that the auxiliary output branch worked very well for the secondary ASR. 
This clearly indicates that the shared layers of the neural network were learned to separate speakers.
In addition, we found out that this 
secondary ASR can be effectively 
incorporated into the proposed method,
which we will explain in the next section.

\subsection{Experiment with dialogue recordings}
\label{sec:dialogue}
Since we confirmed that TS-ASR worked as expected, 
we then conducted experiments for dialogue recordings, 
which were the main target of this study. 

\subsubsection{WER evaluation with oracle non-overlapped speech}

We first evaluated the lower limit of WER for the dialogue recordings 
by using the non-overlapped dialogue recordings (see Section \ref{sec:settings} for recording settings). 
We used the original non-overlapped recordings with the ground-truth segmentation and 
conducted ASR with the clean AM. 
The results are shown in the first line of Table \ref{tab:dialogue}. 
We observed a WER of 19.93\%, which was the lower limit for the recordings in this experiment. 
The WER was worse compared with those for lecture recordings (E1, E2, and E3). 
We observed more substitution errors of backchannels for dialogue recordings, which was very short and difficult to recognize.

\subsubsection{WER evaluation using oracle speaker embeddings}

We then conducted experiments for the mixed monaural dialogue recordings. 
For preprocessing, we separated each dialogue recording into speech segments 
by using simple power-based voice activity detection. 
Note that each segment could contain the speech of two speakers.
We counted a recognized word as correct only when it and the recognized speaker both matched the reference label. 
Since there was ambiguity in the order of speakers in the reference label, 
we calculated the best WER among possible permutations of speakers.

We first conducted an experiment with oracle speaker embeddings to confirm the oracle WER 
for two-speaker mixed recordings. 
The results are shown in the second to fourth rows of Table \ref{tab:dialogue}. 
We extracted the oracle i-vector for each speaker 
by using only the non-overlapped region determined from the ground-truth segmentation.

When we used the clean AM with the oracle speaker embeddings, we observed 
a very poor WER of 94.22\% (\# 2 of Table \ref{tab:dialogue}). 
This was within our expectation because the clean AM was not trained to extract the target speaker's speech.

When we evaluated using the TS-AM 
with embeddings $e_1$ and $e_2$, 
we observed the best oracle WER of 37.96\% (\# 3 of Table \ref{tab:dialogue}). 
This result was the lower limit of WER for two-speaker mixed recordings in this experiment. 
As another application of the TS-AM, we also used the auxiliary output branch of the TS-AM 
with embedding $e_1$ to recognize speaker $2$. 
This result is shown in the fourth row in Table \ref{tab:dialogue}. 
It showed a slightly worse WER of 41.09\% compared with the WER with 
the TS-AM with embeddings $e_1$ and $e_2$ (\# 4 of Table \ref{tab:dialogue}).
This was also within our expectation because the auxiliary branch produced 
a slightly worse result than the main branch according to Table \ref{tab:second}.

\subsubsection{WER evaluation using estimated speaker embeddings with iterative update (proposed method)}

Finally, we evaluated the proposed method 
by starting from the estimated speaker embeddings. 
In this evaluation, we estimated the i-vector for each speech partition divided by 
power-based voice activity detection then applied K-means clustering. 
The number of clusters was set to 2 to be the same as the number of speakers. 
The center of each cluster was used for the initial set of speaker embeddings. 
Note that we denoted the cluster center of the larger cluster as $e_1$ and that of the smaller cluster as $e_2$.

Similar to the comparison of \# 3 and \# 4 of Table \ref{tab:dialogue}, 
we also compared two methods without and with auxiliary output branch. 
The results are shown in the fifth and sixth rows. 
Contrary to the experiment with the oracle speaker embeddings, 
the method using the auxiliary output branch 
with the embedding $e_1$ to recognize the speaker $2$
produced much better WER of 45.54\%
than the method using the main output branch with 
the embedding $e_2$ to recognize the speaker $2$. 
This is because the K-means-clustering-based speaker embedding estimation 
was not sufficient to generate two discriminative embeddings of $e_1$ and $e_2$. 
Considering that embedding $e_2$ was selected as the center of the smaller cluster, 
it would not be as reliable as embedding $e_1$. 
In such a case, using a single embedding $e_1$ with the auxiliary output branch 
is better than using an unreliable embedding $e_2$ with the main output branch. 

We then evaluated the proposed method of applying speaker-embedding estimation and TS-ASR alternately. 
The results are shown in the seventh to ninth rows of Table \ref{tab:dialogue}. 
We observed clear improvement in the WER both for different gender pairs and same gender pairs, and achieved a WER of 40.03\%, which differd by only 2.1\% from the oracle WER of 37.96\% with the oracle speaker embeddings.
Note that we observed better results by the 
proposed method
than that of the method \# 4 of Table \ref{tab:dialogue}
even though the latter method used the speaker embeddings 
obtained from the ground-truth segmentation. 
We believe it was because the ground-truth segmentation 
contained an unignorable amount of silence frames 
that degraded the purity of speaker embeddings, 
while strict exclusion of silence frames was achieved by using the TS-ASR results.

\subsubsection{Evaluation of DERs}

Table \ref{tab:der} lists the DERs of three methods. 
Note that we set 0.25 sec of the no-score collar according to convention,
and calculated DER including overlapped regions.
It is also noted that we regarded silence frames of less than 0.5 sec as speech regions 
for the proposed method because we found the silence information that the ASR produced 
was too strict compared to the ground-truth segmentation developed by human transcribers. 
The first row is a naive method based on the clustering of i-vectors, which was used for the embedding initialization. 
As expected, it produced a very poor DER of 32.37\% due to heavy speech overlaps in the recordings.
 Just by applying TS-ASR, we observed an improvement in the DER to 27.87\%, 
especially for different gender pairs. 
By using the proposed method, the DER further improved to 24.63\%.

To compare with the state-of-the-art  speaker diarization method,
 we also tested the agglomerative hierarchical clustering (AHC)-based method \cite{garcia2017speaker,sell2018diarization}, 
the results of which are shown in the last two rows in Table \ref{tab:der}.
 The i-vector and x-vector extractors were trained using about 3,000 hrs of training data 
consisting of Switchboard-2 (Phase I, II, III), Switchboard Cellular (Part 1, Part2), 
and NIST Speaker Recognition Evaluation (2004, 2005, 2006, 2008) datasets. 
Speaker embeddings were extracted every 0.75 sec, and AHC with probabilistic linear discriminant analysis (PLDA) 
was used to create a speaker cluster. 
Although it is not directly comparable due to the difference in training data, 
we confirmed that our method produced a reasonably good DER although it was trained
much smaller training data.
Our method 
showed a better DER than that of ``i-vector with AHC'' and a DER close to that of ``x-vector with AHC''. 
The proposed method even achieved a better DER than the x-vector-based method 
for the different gender pairs although we used an i-vector trained by much smaller data for the proposed method.

Finally, we discuss the detailed error analysis of the 
DER for the method with speaker embedding updated three times (Table \ref{tab:der-detail}).
We first found that the main source of the DER came from the miss error. 
We also found that the confusion error and false alarm were low even for same gender pairs. 
When applying the proposed method for the different gender pairs, 
almost no confusion and false alarm were produced. 
This means that TS-ASR worked very conservatively, i.e., it tended to output null when 
it was not able to find a reliable word hypothesis. 
From the results in Tables \ref{tab:der} and \ref{tab:der-detail}, 
we can expect further improvement in the DER if
 we apply more discriminative speaker embeddings 
such as d-vector \cite{wang2018speaker,wan2018generalized} and x-vector \cite{garcia2017speaker,snyder2018x}, 
which is one important direction for our future work.

\section{Conclusion}
\label{sec:conclusion}

In this paper, we defined the problem of simultaneous ASR and  speaker diarization,
and proposed an iterative method, 
in which (i) the estimation of speaker embeddings in the recordings 
and (ii) TS-ASR based on the estimated speaker embeddings are alternately executed.
 We evaluated the proposed method by using real dialogue recordings in the 
CSJ
in which the speaker overlap ratio was over 20\%. 
We confirmed that the proposed method significantly 
reduced both the WER and DER. 
Our proposed method with i-vector speaker embeddings 
ultimately 
achieved a WER that differed by only 2.1 \% from the WER of TS-ASR given 
 oracle speaker embeddings. 
Furthermore, our method 
achieved
better DER than that of the conventional clustering-based speaker diarization method based on i-vector. 


There are many directions to enhance this research. 
First, our proposed method should be examined using recordings with more than two speakers. 
Second, the use of more discriminative speaker embeddings, 
such as d-vector \cite{wang2018speaker,wan2018generalized} and x-vector \cite{garcia2017speaker,snyder2018x}, 
will improve performance of both ASR and  speaker diarization. 
Third, the initialization of speaker embeddings should be explored with more advanced speaker diarization techniques \cite{wang2018speaker,fujita2019end,fujita2019end2}.
Finally, advanced ASR techniques, 
such as data augmentation \cite{jaitly2013vocal,kanda2013elastic,ko2015audio,ko2017study}, 
model ensemble \cite{tachioka2013generalized,deng2014ensemble,kanda2017investigation}, improved training criterion \cite{kanda-interspeech2018,weng2019comparison}, will also improve overall performance. 
We will explore these directions for future work.

\bibliographystyle{IEEEbib}
\bibliography{body}

\end{document}